\pdfoutput=1

\documentclass[11pt]{article}

\usepackage[final]{acl}

\usepackage{times}
\usepackage{latexsym}
\usepackage{amsmath}
\usepackage{makecell}
\usepackage{tabularx}
\usepackage{listings}
\usepackage[utf8]{inputenc}

\usepackage[T1]{fontenc}

\usepackage[utf8]{inputenc}

\usepackage{microtype}

\usepackage{tikz}
\usetikzlibrary{positioning}

\usepackage{inconsolata}
\lstdefinelanguage{json}{
    basicstyle=\normalfont\ttfamily,
    showstringspaces=false,
    breaklines=true,
}

\usepackage{graphicx}
\DeclareUnicodeCharacter{2264}{$\leq$}

\title{Mind the Goal: Data-Efficient Goal-Oriented Evaluation of Conversational Agents and Chatbots using Teacher Models}

\author{
  \textbf{Deepak Babu Piskala} \quad
  \textbf{Sharlene Chen} \quad
  \textbf{Udita Patel} \\[2pt]
  \textbf{Parul Kalra} \quad
  \textbf{Rafael Castrillo} \\
  Amazon.com, Seattle, WA, USA
}

\begin{document}
\maketitle
\begin{abstract}
Evaluating the quality of multi-turn chatbot interactions remains challenging, as most existing methods assess interactions at the turn level without addressing whether a user's overarching goal was fulfilled. A ``goal'' here refers to an information need or task, such as asking for policy information or applying for leave. We propose a comprehensive framework for goal-oriented evaluation of multi-agent systems (MAS), introducing the \textbf{Goal Success Rate (GSR)} to measure the percentage of fulfilled goals, and a \textbf{Root Cause of Failure (RCOF)} taxonomy to identify reasons for failure in multi-agent chatbots. Our method segments conversations by user goals and evaluates success using all relevant turns. We present a model-based evaluation system combining teacher LLMs, where domain experts define goals, set quality standards serving as a guidance for the LLMs. The LLMs use ``thinking tokens'' to produce interpretable rationales, enabling \textit{explainable}, \textit{data-efficient} evaluations. In an enterprise setting, we apply our framework to evaluate AIDA, a zero-to-one employee conversational agent system built as a ground-up multi-agent conversational agent, and observe GSR improvement from 63\% to 79\% over six months since its inception. Our framework is generic and offers actionable insights through a detailed defect taxonomy based on analysis of failure points in multi-agent chatbots, diagnosing overall success, identifying key failure modes, and informing system improvements. 

\end{abstract}

\section{Introduction}

\begin{figure}[ht]
\centering
\includegraphics[width=0.9\linewidth]{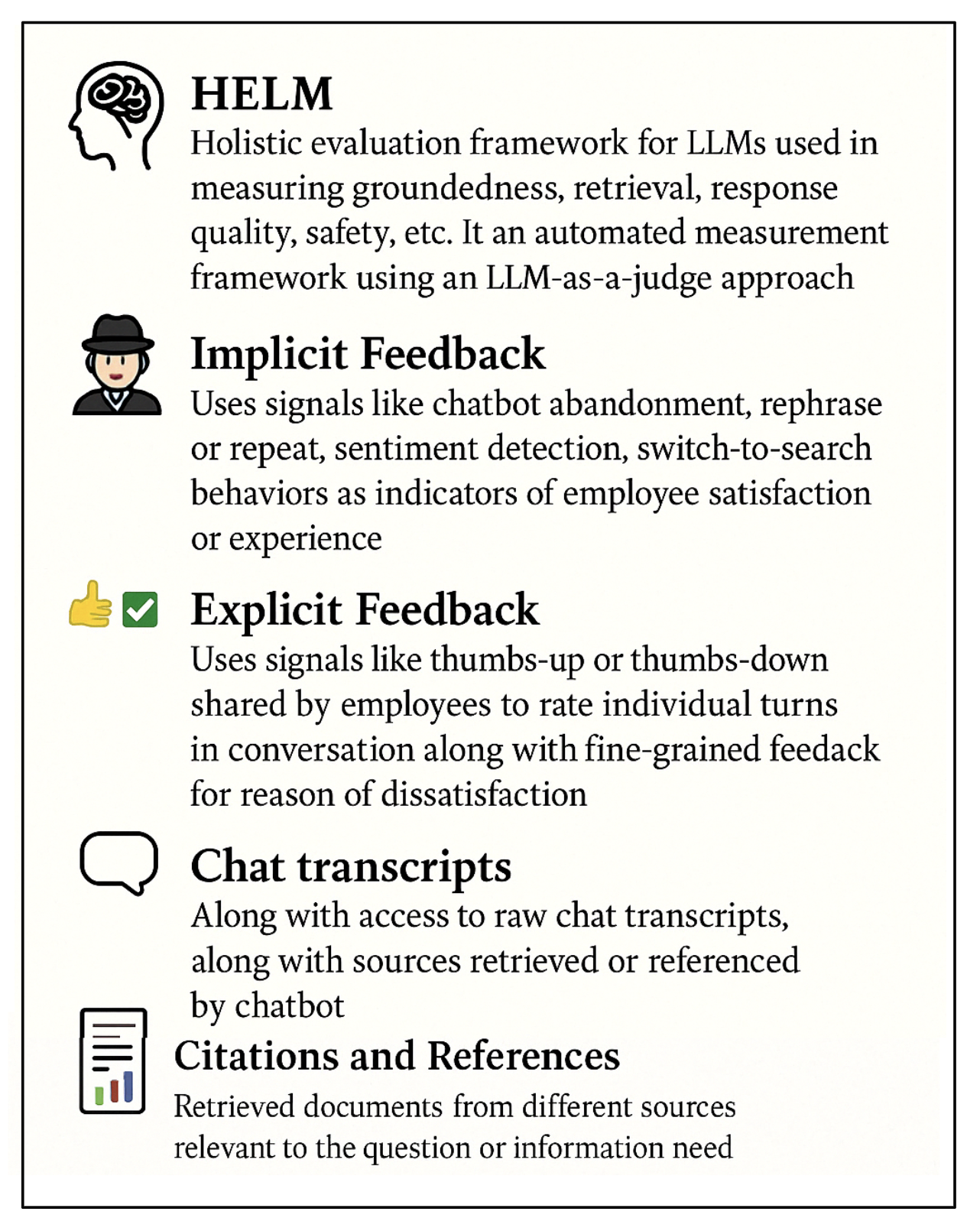}
\caption{Signals used for evaluating chatbot quality}
\label{fig:feedback_signals}
\end{figure}

\begin{figure*}[bt]
  \centering
  \includegraphics[width=\linewidth]{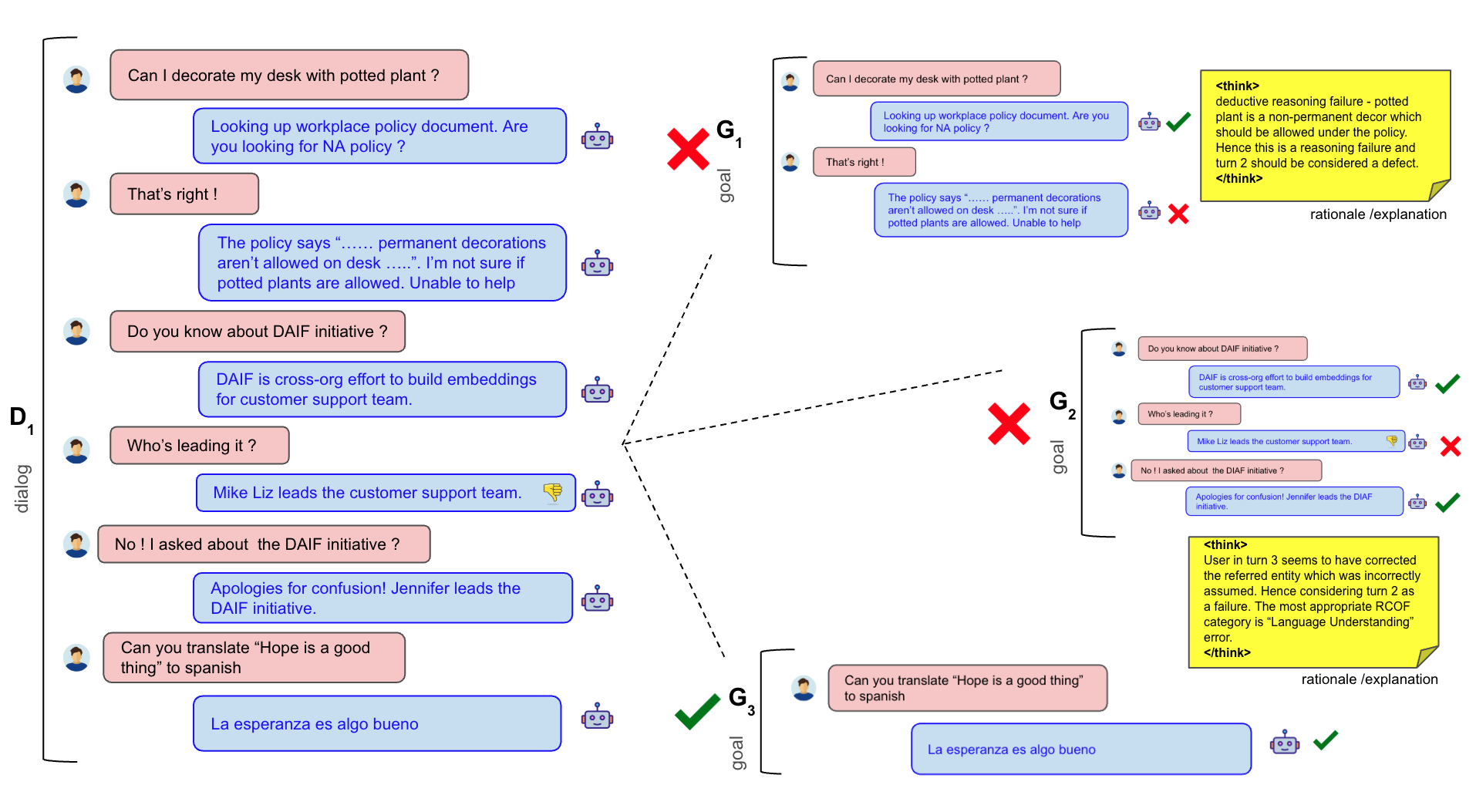}
  \caption{
    \textbf{Goal-oriented breakdown of a multi-turn employee chatbot conversation.}
    The full dialog (left) is segmented into three distinct goals: $G_1$ (policy inquiry), $G_2$ (project clarification), and $G_3$ (translation request). Each goal is independently evaluated for success or failure using goal-level metrics. Turn-level evaluation may suggest high success, but goal-level evaluation reveals that $G_1$ and $G_2$ failed due to reasoning and language understanding errors, respectively. Root Cause of Failure (RCOF) is annotated using structured rationale snippets (right), highlighting the earliest defective turn per goal. 
  }
  \label{fig:goal_breakdown}
\end{figure*}

Modern conversational assistants increasingly adopt agentic LLM architectures, in which a central reasoning model \citep{DeepSeek2025R1,Anthropic2025Claude37,OpenAI2025O3} coordinates tool invocation, external-memory reads/writes, and multi-step planning to accomplish user tasks \citep{Liu2025FoundationAgents,Jacovi2023CoALA}. Such agents learn to call APIs, query databases, or operate forms—often teaching themselves new tool affordances in the wild \citep{Schick2023Toolformer,Qin2024ToolSurvey}. To support long-horizon dialogues and personalization, they maintain dedicated episodic or vector memories that can be dynamically written and retrieved \citep{Salama2025MemInsight,Du2023LongMem}. Sophisticated reasoning routines (e.g., chain-of-thought or program-of-thought prompts) enable the agent to decompose queries, condition tool use on intermediate results, and verify answers before replying \citep{Wei2022CoT,Chen2022PoT}. At enterprise scale, multiple specialized agents—HR, IT, legal, analytics—often collaborate through emerging interoperability protocols such as Anthropic’s Model Context Protocol (MCP) and Google’s Agent-to-Agent (A2A) standard, forming loosely coupled multi-agent systems \citep{Hou2025MCP,Khandelwal2025A2A}. While this ecosystem unlocks powerful organizational workflows, every new memory layer, tool wrapper, or inter-agent message channel compounds the risk of subtle cascading failures, underscoring the need for robust, interpretable evaluation.

Chatbots and conversational assistants are increasingly used to handle complex information-seeking and action-taking dialogues. Ensuring high-quality interactions is critical, especially as users may ask follow-up questions or rephrase queries until their goal is met. However, evaluating chatbot quality in a meaningful way remains challenging. Today, most chatbot evaluations focus on individual turn-level metrics (each user query and the bot's response), such as response relevance \cite{patel2025}. While these metrics provide useful signals, they often fall short of capturing the full picture of user satisfaction --- in particular, whether the user's \emph{underlying goal} was eventually achieved across the entire conversation. 

There is a need for a more holistic evaluation framework that moves beyond isolated turns to assess the success of a conversation as a whole. Current systems lack a unified view to diagnose end-to-end conversational success or identify where in a multi-turn exchange the assistant failed to meet the user's needs. For example, an enterprise chatbot may involve components for retrieval, language understanding, and external tool or database calls; a failure in any one component can cause the conversation to derail.

In practice, multi-turn dialogs are common and particularly prone to failures. In our analysis of an enterprise conversational assistant called AIDA, we found that about 39\% of dialogs involve multiple turns, and these multi-turn dialogs contribute to a disproportionate share of user frustration. For instance, multi-turn sessions exhibited a negative feedback rate (e.g. user explicitly indicating dissatisfaction) roughly three times higher than single-turn sessions (2.65\% vs 0.9\%).

We propose a goal-oriented evaluation framework for chatbots that segments each dialog into user-defined \textbf{goals} and measures a strict \textit{Goal Success Rate (GSR)} — a goal is marked successful only if \emph{all} its turns are error-free. To make this metric actionable, we also introduce a \textit{Root Cause of Failure (RCOF)} taxonomy that attributes each failed goal to a predefined error category, enabling developers to pinpoint dominant failure modes. In summary, our contributions include:
\begin{itemize}
    \item A general \textbf{goal-oriented evaluation framework} for multi-turn conversations, which segments dialogs by user goals and evaluates success at the goal level.
    \item Definition of the \textbf{Goal Success Rate (GSR)} metric to quantify the fraction of user goals that are satisfied, and a \textbf{Root Cause of Failure (RCOF)} taxonomy to categorize and explain failed goals.
    \item A \textbf{model-based implementation} using a large language model as a ``teacher'' to assist in labeling goals and turn outcomes, demonstrating how GSR and RCOF can be computed on real chatbot logs.
    \item Empirical analysis on enterprise chatbot conversations, showing that the framework captures holistic quality signals (e.g. lower GSR for multi-turn queries, identification of top failure reasons) and discussing how these insights drive system improvements.
\end{itemize}

\section{Related Work}
Evaluating open-domain and task-oriented dialog systems has been an active research area. Traditional evaluation metrics for chatbots often operate at the \emph{utterance} level. For example, automated metrics such as BLEU~\cite{papineni2002bleu} and ROUGE~\cite{lin2004rouge} compare generated responses to reference texts, while embedding-based scores like BERTScore~\cite{zhang2019bertscore} aim to capture deeper semantic similarity. More recently, large language models themselves have been employed as judges—rating responses against conversational guidelines or rubrics (e.g.\ G-Eval~\cite{liu2023geval} and MT-Bench evaluations~\cite{zheng2023judging}). However, these turn-level metrics do not directly indicate whether the user’s \emph{broader goal} was accomplished.

In task-oriented dialogue systems, it is common to measure \textit{task success} or \textit{goal completion rate} \cite{Lu2020}. These metrics typically require a predefined goal (such as booking a restaurant or answering a specific question) and often rely on comparing the dialog outcome to a known target or using user self-reports. Our work is inspired by task success metrics \cite{Lu2020} but extends the idea to more open-ended, user-driven conversations where the goals must be inferred and may evolve.

Another line of related work focuses on user satisfaction \cite{FU202214} and engagement as metrics for dialog quality. For example, systems have been evaluated based on user ratings, re-engagement rates, or negative feedback signals (e.g. explicit thumbs-down clicks) \cite{hancock2019learningdialoguedeploymentfeed, Mehri2020USR}. These signals provide valuable supervision for quality, and our framework could integrate them (e.g. as features or validation for goal success labels). Nonetheless, they are often sparse and do not give detailed reasons for failure.

In terms of understanding failure modes, prior work in error analysis of conversational systems and virtual assistants has introduced taxonomies of errors (for instance, distinguishing between interpretation errors vs. knowledge retrieval errors) \cite{ram2018conversationalaisciencealexa}. Our RCOF taxonomy is aligned with these ideas, but tailored to the enterprise chatbot scenario and designed to be used in an automated evaluation pipeline. Lastly, we leverage large pre-trained language models to assist evaluation, which relates to the growing trend of using AI models as proxy evaluators or ``rubric generators'' for AI outputs \cite{zhang2022finedeval, mao2023gpteval}. Our use of a teacher model demonstrates how such models can help segment and label conversation quality at scale, while a human-in-the-loop ensures the reliability of these labels for building goal-level metrics. Lu et al.~\cite{lu2020efficient} propose a BERT-based span prediction model to optimize the identification of relevant contextual utterances in task-oriented dialogues. This approach enhances the calculation of Goal Success Rate (GSR) by accurately segmenting user goals, thereby reducing labeling waste and potential bias in evaluation metrics.

\section{Anatomy of a Conversation}
To effectively evaluate chatbot interactions, we begin by defining the fundamental units of analysis: \textbf{session}, \textbf{goal}, and \textbf{turn}. These definitions allow us to formally model dialogue structure and success.

\begin{itemize}
    \item A \textbf{session} ($S$) refers to a full interaction between a user and the chatbot, typically bounded by a timeout or user exit (also called dialog).
    \item Each session consists of one or more \textbf{goals} ($G_i$), where a goal represents a coherent user intent or information need (e.g., “Where can I submit expenses?”).
    \item A \textbf{goal} is realized as a sequence of one or more \textbf{turns} ($T_j$), with each turn comprising a user query $q_j$ and the chatbot response $r_j$.
\end{itemize}

A goal is marked \textit{successful} if all its turns are successful\footnote{Alternate formulations include assigning fractional credit based on the number of successful turns, or defining goal success based on the final turn alone. We discuss these in the Future Directions section.}—i.e., the bot provided correct and helpful responses. If any turn within a goal fails, the entire goal is considered failed. This strict criteria ensures high fidelity to user experience.

\begin{figure}[ht]
    \centering
    \includegraphics[width=0.9\linewidth]{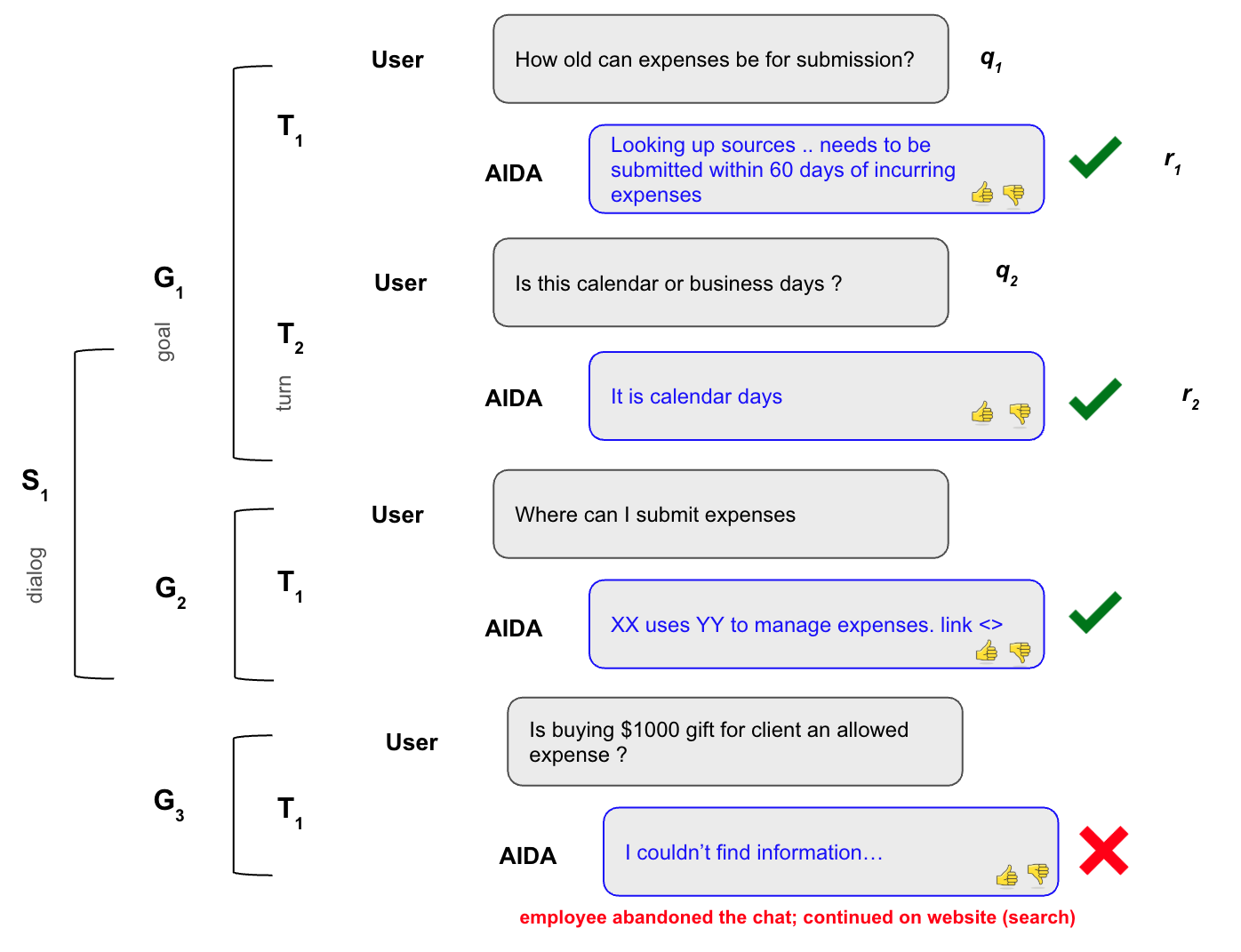}
    \caption{Anatomy of a chatbot session: Each dialog consists of multiple goals ($G_i$), where each goal comprises one or more turns ($T_j$), formed by a query-response pair ($q_j$, $r_j$). Goal $G_3$ is marked failed due to information unavailability.}
    \label{fig:anatomy}
\end{figure}

Figure~\ref{fig:anatomy} illustrates an example session $S_1$ comprising three goals ($G_1$, $G_2$, and $G_3$), each containing one or more turns. The first two goals are successful, while the third fails due to missing knowledge, prompting the user to abandon the chat.

\section{Framework for Goal-Oriented Evaluation}
Our framework, which we refer to as a Conversational Intelligence Model (CIM) for chatbot evaluation, consists of three main components: (1) \textbf{Goal Segmentation}, (2) \textbf{Goal Success evaluation (GSR)}, and (3) \textbf{Root Cause of Failure (RCOF) attribution}. We describe each in turn.

\subsection{Goal Segmentation}
A conversational session (dialog) is a sequence of turns, where each turn typically consists of a user’s message (question) and the chatbot’s response. In goal segmentation, the aim is to identify boundaries between distinct \emph{goals} within a dialog. Intuitively, a new goal begins when the user asks something that constitutes a new or different information need, as opposed to continuing or clarifying the previous question. 

In formal terms, for each turn $T_j$ in a dialog (for $j=1,2,\ldots,N$), we predict a label $\textit{is\_new\_goal}(T_j) \in \{\textit{yes}, \textit{no}\}$ indicating whether $T_j$ starts a new goal. The first turn of a dialog is always a new goal by definition. A label of \textit{yes} means the user’s utterance in turn $T_j$ is considered the start of a new goal (information need), whereas \textit{no} means the turn is continuing the previous goal. By applying this segmentation, a single dialog of $N$ turns can be divided into one or more goals $G_1, G_2, \dots, G_K$, where $K$ is the number of turns labeled as starting a new goal. Each goal $G_k$ consists of a contiguous sequence of turns addressing a particular query or task.

Goal segmentation can be seen as a classification task at the turn level. Features for this task can include lexical cues (e.g. the user asks an unrelated question indicating a topic switch), contextual cues (e.g. the user explicitly says "Now I have another question..."), or even temporal gaps (if a conversation resumes after a long pause, it might indicate a new goal). In our implementation, we utilize a large language model to examine the conversation and predict these boundaries, as described in \ref{sec:methodology}. Accurate segmentation is important, since it directly affects the granularity at which success is measured.

\begin{figure*}[tb]
  \centering
    \includegraphics[width=\linewidth]{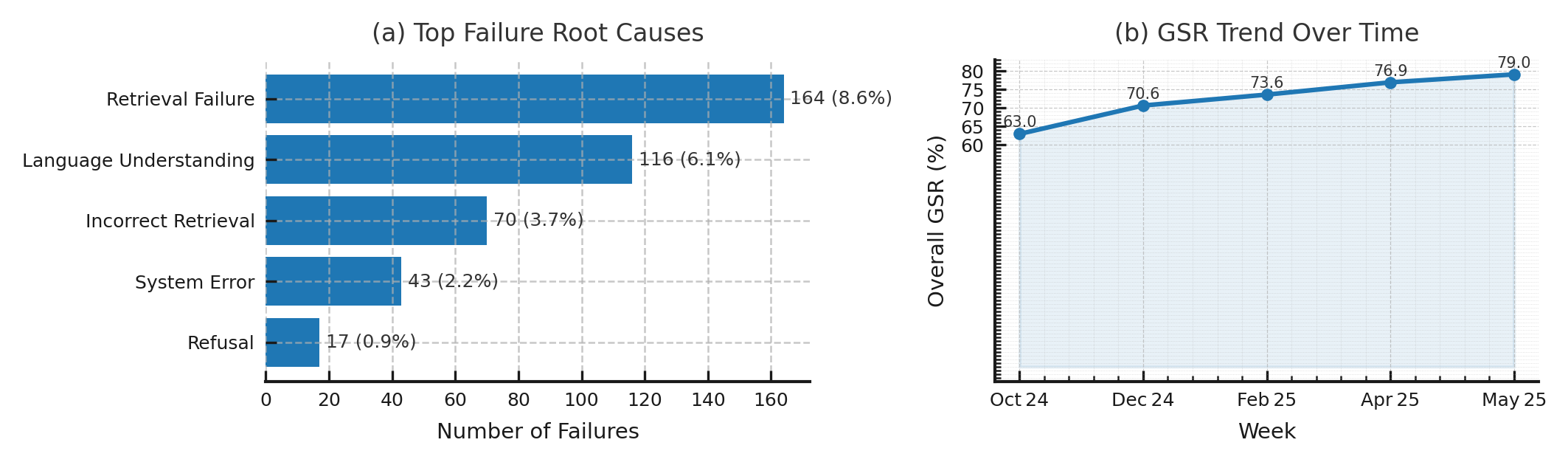}
    \label{fig:fail-roots}
  \caption{%
    \textbf{(a)} Distribution of the top five chatbot failure root causes, annotated with both absolute counts and percentages. 
    \textbf{(b)} Trend of the overall Goal-Success Rate (GSR) from Oct ’24 through May ’25, showing a steady improvement.%
  }
  \label{fig:combined}
\end{figure*}

\subsection{Goal Success Rate (GSR)}
Once goals are identified in a conversation, we evaluate the quality of each goal by examining the turns within it. We define a \textbf{goal} as \emph{successful} if \textbf{every turn in that goal is successful}. Conversely, if any turn in the goal was a failure (e.g. the chatbot's answer was incorrect, irrelevant, or otherwise unsatisfactory for that turn), then the entire goal is labeled as \emph{failed}. This definition is deliberately strict: even if the user eventually gets the answer after rephrasing their question in a later turn, we consider the goal to have failed because the user had to work through a failed response along the way. In other words, the assistant must get it right on the first attempt \emph{and} continue to be correct for all follow-ups to count as a success under this metric. This high standard ensures that our evaluation emphasizes truly seamless interactions.

Formally, let each goal $G_k$ consist of turns $T_{s_k}, T_{s_k+1}, \dots, T_{e_k}$, where $T_{s_k}$ is the first turn of goal $k$ and $T_{e_k}$ is the last turn (right before the next goal starts or the dialog ends). We have a function $\textit{quality}(T_j)$ which outputs \textit{success} or \textit{failure} for each turn (how to determine turn success is discussed later). Then the quality of goal $G_k$ is:

\begin{equation*}
\begin{split}
\textit{GoalQuality}(G_k)\\
=\quad
\begin{cases}
  \textit{success}, & \text{if } \forall\,T_j \in G_k,\;\textit{quality}(T_j)=\textit{success},\\
  \textit{failure}, & \text{otherwise}.
\end{cases}
\end{split}
\end{equation*}

The \textbf{Goal Success Rate (GSR)} is defined as the fraction of goals in a dataset that are successful. If we have $K$ goals in total (across some set of dialogs), and we use an indicator function $1[\cdot]$ that is 1 if a goal is successful and 0 if failed, then:
\[
\text{GSR} = \frac{1}{K}\sum_{k=1}^{K} 1[\textit{goalQuality}(G_k) = \textit{success}] \times 100\%\,.
\]

\subsection{Root Cause of Failure (RCOF) Taxonomy}
While GSR provides a high-level success metric, it does not explain \textit{why} goals fail. To enable actionable insights, we introduce a taxonomy of seven error categories—termed \textbf{Root Cause of Failure (RCOF)}—each corresponding to a distinct breakdown in chatbot behavior: \textbf{E1} Language Understanding Failure, \textbf{E2} Refusal to Answer, \textbf{E3} Incorrect Retrieval, \textbf{E4} Retrieval Failure, \textbf{E5} System Error, \textbf{E6} Incorrect Routing, and \textbf{E7} Out-of-Domain or Unsupported Query. Each failed turn is annotated with one of these codes, and we assign a goal’s RCOF based on the earliest failed turn, under the assumption that initial breakdowns are most disruptive. This framing helps prioritize debugging and aligns naturally with engineering signals (e.g., E3 with low coverage, E5 with timeouts), making the taxonomy interpretable and actionable. A detailed description of each RCOF category is provided in \textbf{Appendix~\ref{sec:rcof}}.

\section{Data}
\label{sec:data}
We evaluate our GSR framework on real-world interaction logs from \textbf{AIDA}, an enterprise-grade virtual assistant deployed across desktop and mobile to help employees with workplace queries spanning HR, IT, wiki, expenses, and internal tools. The dataset comprises approximately \textasciitilde10{,}000 multi-turn conversations collected over 30 days, where each session may embed multiple user \textbf{goals}—ranging from informational queries to action-oriented tasks like leave applications or meeting room bookings. Each turn is annotated with rich \textbf{implicit signals} (rephrases, abandonments, search fallback), \textbf{explicit feedback} (likes, thumbs up/down), and \textbf{metadata} such as device type, timestamp, and retrieved citations. A detailed breakdown of dataset composition and feedback signals is provided in \textbf{Appendix~\ref{sec:dataset}}.

\section{Methodology}
\label{sec:methodology}

To generate ground-truth annotations for evaluating chatbot quality, we adopt a human-in-the-loop (HITL) pipeline that combines expert-defined SOPs with LLM-based multi-teacher supervision. The methodology applies across all stages of evaluation: goal segmentation, success classification (GSR), and root cause attribution (RCOF).

\begin{figure}[ht]
\centering
\includegraphics[width=0.95\linewidth]{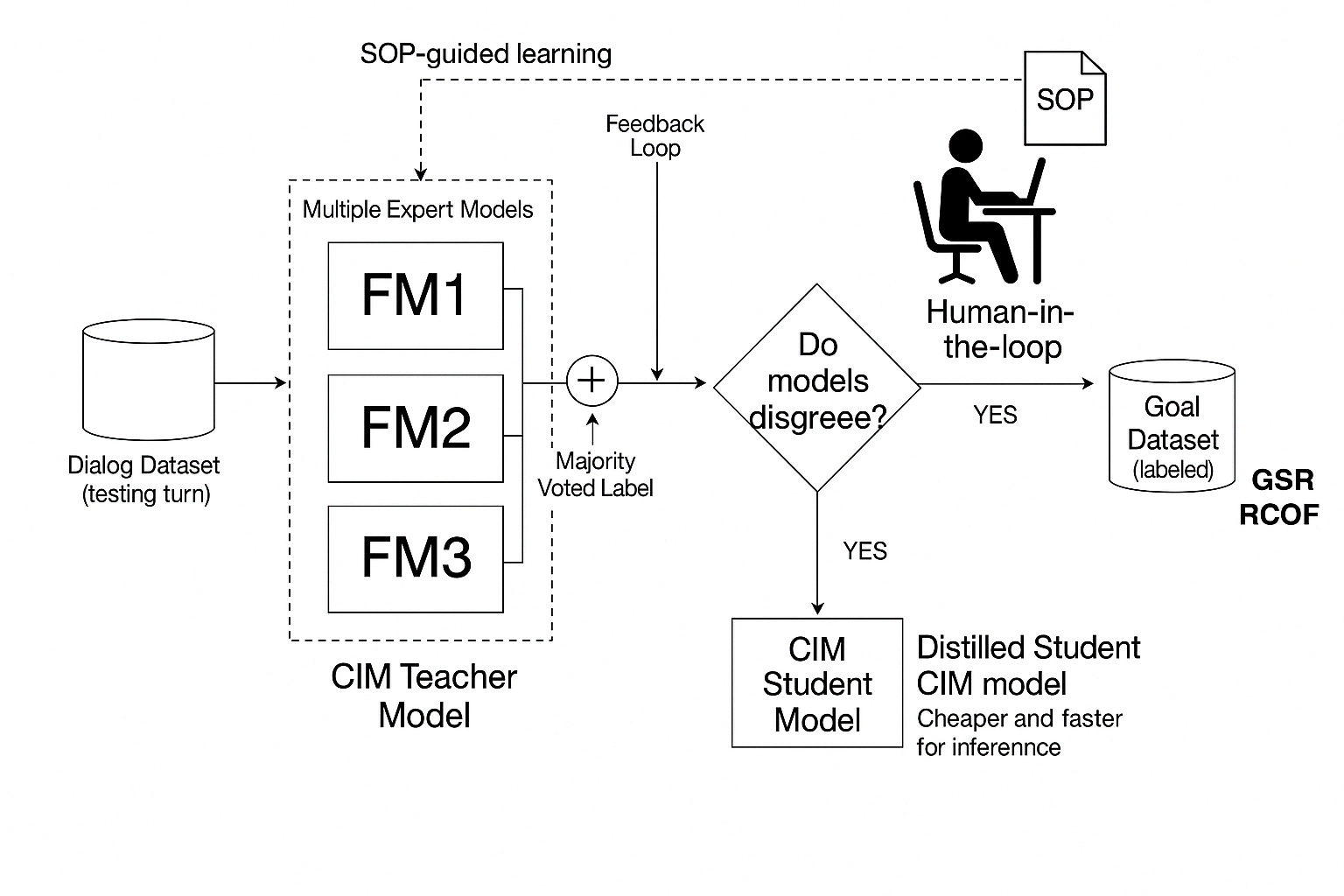}
\caption{HITL evaluation pipeline: AIDA conversations are processed by multiple expert models using Chain-of-Thought prompts. Majority-voted outputs are accepted as labels. Disagreements are escalated to human experts guided by SOPs. }
\label{fig:method_architecture}
\end{figure}

Figure~\ref{fig:method_architecture} illustrates our end-to-end setup. We begin with normalized event logs collected from AIDA, an enterprise chatbot. These logs are preprocessed into a linked dialog dataset, grouping message turns into coherent multi-turn conversations. We oversample longer sessions to ensure coverage of complex goals.

We then sample $N$ conversations and evaluate each using a set of three independently prompted foundation models (FMs), referred to as expert teacher models—examples include Claude Sonnet, Claude Haiku \cite{anthropic_haiku_2024}, GPT-4 \cite{openai_gpt4_2023}, and LLaMA-4 \cite{meta_llama4_2025}. Each model is invoked using Chain-of-Thought (CoT) prompting, wherein the model is instructed to use explicit reasoning tags (`<think> ... </think>`) before outputting its final quality judgment or label. This promotes reflective system-2 style thinking and enables richer, interpretable rationales.

Once each expert provides its opinion on a given goal (e.g., whether it was fulfilled or which failure label applies), we aggregate the responses via majority voting. If two or more teacher models agree on a label, we accept the annotation as ground truth. In cases where all three models disagree, we mark the goal as \textit{ambiguous} and escalate to human annotators. These experts refer to business-specific SOPs to resolve such edge cases and refine definitions of quality. Over time, this feedback loop refines model behavior and improves inter-model consistency.

Once the labeled goal dataset is produced, we optionally distill the teacher ensemble into a lightweight student model for efficient real-time and offline inference. The student, trained on the voted labels, mimics teacher decisions at a lower cost.

\section{Results}
\label{sec:results}
We evaluated the GSR framework on a stratified sample of approximately 10,000 dialogs from \textbf{AIDA}, as described in Section~\ref{sec:data}. The overall Goal Success Rate (GSR) was 78\%, indicating that a majority of user goals were successfully resolved end-to-end without any defective turns. However, the GSR dropped to 66\% for multi-turn goals (those spanning two or more turns), revealing that conversational complexity significantly increases the risk of failure. A breakdown of goal failures using our RCOF taxonomy is shown in Table~\ref{tab:gsr}, where the top three failure types were retrieval failures (39\%), language understanding errors (27\%), and incorrect retrievals (16\%). This confirms that while AIDA handles single-turn factoid queries well, it is more prone to error in nuanced or multi-step interactions where accurate retrieval and contextual comprehension are critical.

\begin{table}[h]
\centering
\footnotesize
\setlength{\tabcolsep}{4pt}
\begin{tabular}{|p{3.3cm}|c|c|}
\hline
\textbf{Metric} & \textbf{Count} & \makecell{\textbf{\%} \\ \textbf{of Goals}} \\
\hline
Total Goals (sample) & 1915 & 100\% \\
Successful Goals & 1488 & 77.7\% \\
Failed Goals & 427 & 22.3\% \\
\hline
\multicolumn{3}{|l|}{\textit{Top failure root causes:}} \\
\hline
\makecell[l]{Retrieval Failure (E4)} & 164 & 8.6\% \\
\makecell[l]{Language Understanding (E1)} & 116 & 6.1\% \\
\makecell[l]{Incorrect Retrieval (E3)} & 70 & 3.7\% \\
\makecell[l]{System Error (E5)} & 43 & 2.2\% \\
\makecell[l]{Refusal (E2)} & 17 & 0.9\% \\
\hline
\end{tabular}
\caption{Goal Success Rate and failure breakdown for a sample of AIDA chatbot dialogs. The top section shows overall GSR, and the bottom lists top root causes of failed goals.}
\label{tab:gsr}
\end{table}

Over a three-month period, AIDA evolved from a basic retrieval-augmented system to an agentic LLM-powered assistant capable of reasoning, invoking tools, and managing contextual queries. As shown in Figure~\ref{fig:combined}, we observed a steady improvement in goal completion rates—from 64\% in February to 78\% by April. This growth was not driven by prompt tuning or fallback rules, but rather by launching new capabilities: improved source integration, routing mechanisms, upgraded models with better language reasoning, and more flexible agentic behaviors (e.g., issuing clarification questions or synthesizing multi-source answers). Notably, the GSR for multi-turn goals rose by 12 points, demonstrating the practical utility of our evaluation framework in guiding and validating iterative system improvements. We assess teacher model reliability by comparing its labels against expert human annotations; detailed agreement statistics and analysis are provided in \textbf{Appendix~\ref{sec:human-agreement}}.

\section*{Limitations}

While our proposed goal-oriented evaluation framework offers a structured and scalable way to assess chatbot quality, it has certain limitations.

First, our evaluation methodology is best suited for task-oriented and information-seeking dialogs with clear success criteria. In open-ended scenarios such as summarizing documents or composing emails, quality becomes highly subjective and user-dependent. Behavioral signals like thumbs-downs, chat termination, or repeated clarifications can offer implicit supervision, but such signals are sparse and often unavailable in real-world logs.

Second, we assume that each goal corresponds to a contiguous sequence of turns. However, in complex or compound dialogs, user goals may span non-consecutive turns or interleave with others (e.g., returning to a prior topic). Our current segmentation method does not support such dependencies. Future work could explore modeling dialog goals as graph structures to capture cross-references, co-references, and interleaved subgoals more accurately.

Third, our current evaluation may understate hallucinations—cases where the assistant generates fluent but factually incorrect information. Without external verification or user behavior signals to flag discrepancies, such errors may go undetected. Addressing hallucination detection remains a broader challenge, especially in enterprise scenarios where accurate grounding in internal knowledge bases is critical.

\bibliography{custom}

\clearpage            
\appendix             
\section*{Appendix}   
\label{sec:appendix}

\section{Dataset Composition}
\label{sec:dataset}
We evaluate our GSR framework using real-world interaction logs from an enterprise-grade chatbot, called AIDA deployed to assist employees with workplace-related queries. The chatbot serves as a virtual assistant capable of addressing a wide variety of topics including HR policies, IT troubleshooting, expense reimbursement, time-off requests, and access to internal tools or documentation. The assistant operates in natural language and is available via chat platforms on desktop and mobile, used internally by the organization. In addition to answering informational queries, AIDA can handle action-oriented goals such as applying for sick leave on behalf of an employee or booking a meeting room.

Our dataset comprises $\sim$10{,}000 multi-turn conversations collected over a 30-day window. Each conversation represents a \textbf{session}, which may contain multiple \textbf{goals}, where a goal corresponds to a specific user intent or task (e.g., checking leave balance, updating benefits, submitting expenses). The data includes:

\begin{itemize}
    \item \textbf{User utterances and bot responses}: Complete conversational transcripts segmented into turns.
    \item \textbf{Implicit feedback signals}: Indicators such as query rephrasing, abandonment, switches to search, and delayed user responses.
    \item \textbf{Explicit feedback}: Thumbs up/down on responses, likes, and internal reshares.
    \item \textbf{Citations and References}: Includes RAG articles used for answering the query, tool outputs, etc. along with metadata of conversationa like time, device(mobile/desktop) and user attributes.  
\end{itemize}

\section{Detailed RCOF Definitions}  
\label{sec:rcof} 
While GSR provides a single-number summary of success, it does not explain \textit{why} goals failed. For actionable insights, we need to analyze the failures in more detail. We introduce a predefined taxonomy of error categories to attribute each failed goal to a \textit{root cause of failure (RCOF)}. Each failure category corresponds to a general type of breakdown that can occur in a chatbot's handling of a query.

Drawing from common issues in information-seeking and action-taking dialogues, we define seven distinct root cause categories (which we label E1 through E7 for convenience):
\begin{itemize}
    \item \textbf{Language Understanding Failure (E1)}: The assistant misunderstood the user's request or context, leading to an irrelevant or incorrect answer. For example, the user says "cancel my request," and the bot misinterprets "cancel" in the wrong context.
    \item \textbf{Refusal to Answer (E2)}: The assistant inappropriately refused to answer the question (or gave a safe completion) even though it should have been able to help. In other words, no disallowed content was present, but the bot still responded with a refusal.
    \item \textbf{Incorrect Retrieval (E3)}: The assistant retrieved the wrong informational content. This is specific to systems that use a retrieval-augmented generation (RAG) or knowledge base: the bot did fetch some documents or data, but those turned out to not contain the answer needed (so the answer was inevitably wrong or incomplete).
    \item \textbf{Retrieval Failure (E4)}: The assistant failed to retrieve any relevant information when it should have. For instance, the user asked a factual question answerable from a knowledge base, but the system returned no results (perhaps due to a search/query failure).
    \item \textbf{System Error (E5)}: A technical issue prevented a correct answer. This could include the response getting cut off (e.g. a timeout or the generation stopping mid-sentence) or an integration failure. Essentially, the system could not produce a proper answer due to an error or glitch.
    \item \textbf{Incorrect Routing (E6)}: The user's query was routed to the wrong domain or module of the assistant. In enterprise assistants that orchestrate multiple bots or skill routes, a question might be answered by an inappropriate knowledge category (for example, a question meant for an HR database was mistakenly handled by a general FAQ bot), leading to a faulty answer.
    \item \textbf{Out-of-Domain or Unsupported Query (E7)}: The user’s request is outside the scope of what the assistant is designed to handle (e.g. asking a legal question to an IT support bot), or involves capabilities it doesn't have (like requesting a translation if that's not supported). In such cases, failure is expected because the question is invalid for the system.
\end{itemize}

Each failed turn in a conversation is annotated with an RCOF code (E1--E7) to indicate the failure type. Since a single goal may contain multiple failed turns, we define the \textbf{root cause of a failed goal} as the error category of its \emph{earliest failed turn}. This heuristic assumes that the initial breakdown is the most influential in derailing the goal and helps focus analysis on the first error rather than compounding effects. For example, if the assistant misunderstood the question (E1) early on, we attribute the goal’s failure to language understanding—even if a retrieval failure occurred later. In other words, within a failed goal $G_k$, let $T_j$ be the first turn (in chronological order) marked as failure; the RCOF label assigned to $T_j$ becomes the root cause for $G_k$. This approach guides debugging toward root issues rather than symptoms. Moreover, RCOF categories can be aligned with internal system metrics (e.g., low source coverage may indicate E3, or frequent fallbacks may reflect E5), making the taxonomy both interpretable and actionable for engineering teams.

\section{Human–LLM Agreement}
\label{sec:human-agreement}
\textbf{Agreement with Human Annotators}: To assess the reliability of our teacher model ensemble, we conducted a comparison against expert human annotators using the same SOP guidelines provided to the LLMs. We found that human reviewers agreed with model-generated labels in approximately \textbf{75\%} of cases. In the remaining \textbf{25\%}, there was at least one point of disagreement across the three annotation tasks: \emph{goal segmentation}, \emph{turn-level quality}, and \emph{root cause attribution (RCOF)}.

When analyzing task-specific agreement, we observed that only \textbf{13\%} of dialogs showed disagreement between humans and LLMs for either goal segmentation or turn quality evaluation. However, for RCOF attribution, disagreement rose to \textbf{17\%} of cases. We attribute these gaps to \emph{ambiguity in the SOP definitions}, where both humans and models encountered unclear guidance for edge cases or subjective interpretations. We expect to reduce such discrepancies to below \textbf{5\%} through \emph{closed-loop train–evaluate cycle} that integrates human feedback into the teacher model prompting strategy.


\section{LLM Prompt Template}
\label{sec:prompt}

\begin{lstlisting}[language={},basicstyle=\ttfamily\footnotesize,
                   columns=fullflexible,frame=single,breaklines=true]
system_prompt = "You are a helpful AI assistant. You will act as a judge to evaluate quality of employee experience chatbot."

output_format = """
{
  dialog_id: xx,
  turns: [
    {turn_number: 1, is_new_goal: yes/no, quality: success/failure, rcof: E1-E7 | null},
    {turn_number: 2, is_new_goal: yes/no, quality: success/failure, rcof: E1-E7 | null},
    ...
  ]
}

where
  is_new_goal \in {yes,no}               # compare adjacent user turns
  quality     \in {success,failure}      # based on response + follow-ups
  rcof        \in {E1-E7} if failure else null

RCOF codes
  E1 Incorrect Sources        - irrelevant docs retrieved
  E2 Retrieval Failure        - no docs retrieved
  E3 Refusal to Answer        - unwarranted refusal
  E4 Language Understanding    - misinterprets question
  E5 System Error             - blank / truncated response
  E6 Incorrect Routing        - wrong domain/department
  E7 Out-of-Domain Query      - capability not supported
"""

template = """
{system_prompt}
You are provided with a dialog from an employee chatbot. 
Output the JSON for every turn, reasoning inside <think>...</think> tags 
but printing *only* the JSON.

output format:
{output_format}

input:
{question}
"""
\end{lstlisting}

\section{Custom JSON Schema}
\label{sec:json-schema}

\noindent\textbf{Schema overview.}  
We store every annotated dialog as a JSON object with a unique \texttt{dialog\_id} and an array of per-turn records.  
Listing~\ref{lst:schema} shows the structure (placeholders \texttt{<…>} indicate value types).

\begin{lstlisting}[language=json,basicstyle=\ttfamily\footnotesize,
                   columns=fullflexible,frame=single,
                   caption={Abstract JSON schema for dialog annotations used in our pipeline},
                   label={lst:schema}]
{
  "dialog_id": <string>,          // UUID for the dialog
  "turns": [
    {
      "turn_number": <int>,       
      "user_msg": <string>,       
      "response": <string>,       
      "source_urls": <string[]>,  
      "source_names": <string[]>, 
      "source_snippets": <string[]> 
    },
    ...
  ]
}
\end{lstlisting}

\paragraph{Field descriptions.}
\begin{itemize}
  \item \textbf{\texttt{dialog\_id}} – Globally unique identifier (UUID v4) used to join logs and annotations.
  \item \textbf{\texttt{turn\_number}} – Sequential index starting at 1; enables mapping annotations back to raw logs.
  \item \textbf{\texttt{user\_msg}} / \textbf{\texttt{response}} – Raw text of the employee’s utterance and the chatbot’s reply.
  \item \textbf{\texttt{source\_urls}} – List of URLs or internal document IDs returned by the retrieval component.
  \item \textbf{\texttt{source\_names}} – Optional human-friendly titles corresponding to each URL (may be empty).
  \item \textbf{\texttt{source\_snippets}} – Evidence snippets (≤256 chars each) extracted from the retrieved sources and shown to the user.
\end{itemize}

This schema underpins both the teacher-model annotation pipeline and downstream analytics, ensuring that every evaluation label can be traced back to its conversational and knowledge context.

\end{document}